\documentclass[runningheads]{llncs}

\usepackage[mobile]{eccv}

\usepackage{eccvabbrv}
\usepackage{bm}
\usepackage{graphicx}
\usepackage{booktabs}
\usepackage{comment}

\usepackage[accsupp]{axessibility}  

\usepackage{hyperref}

\usepackage{orcidlink}
\usepackage{marvosym}
\newcommand{\DynMF}{DynMF}

\begin{document}

\title{\DynMF: Neural Motion Factorization for Real-time Dynamic View Synthesis with 3D Gaussian Splatting} 


\author{Agelos Kratimenos \orcidlink{0009-0006-7179-7944}\inst{1}\and
Jiahui Lei \orcidlink{0009-0008-1108-7953}\inst{1\dag}  \and
Kostas Daniilidis \orcidlink{0000-0003-0498-0758}\inst{1,2} }
\authorrunning{A. Kratimenos et al.}

\institute{University of Pennsylvania, Philadelphia, PA, USA \and
Archimedes, Athena RC}

\titlerunning{DynMF: Dynamic Neural Motion Factorization}

\maketitle
\begin{abstract}
  Accurately and efficiently modeling dynamic scenes and motions is considered so challenging a task due to temporal dynamics and motion complexity. To address these challenges, we propose \DynMF, a compact and efficient representation that decomposes a dynamic scene into a few neural trajectories. We argue that the per-point motions of a dynamic scene can be decomposed into a small set of explicit or learned trajectories. Our carefully designed neural framework consisting of a tiny set of learned basis queried only in time allows for rendering speed similar to 3D Gaussian Splatting, surpassing 120 FPS, while at the same time, requiring only double the storage compared to static scenes. Our neural representation adequately constrains the inherently underconstrained motion field of a dynamic scene leading to effective and fast optimization. This is done by biding each point to motion coefficients that enforce the per-point sharing of basis trajectories. By carefully applying a sparsity loss to the motion coefficients, we are able to disentangle the motions that comprise the scene, independently control them, and generate novel motion combinations that have never been seen before. We can reach state-of-the-art render quality within just 5 minutes of training and in less than half an hour, we can synthesize novel views of dynamic scenes with superior photorealistic quality. Our representation is interpretable, efficient, and expressive enough to offer real-time view synthesis of complex dynamic scene motions, in monocular and multi-view scenarios. 
\end{abstract}

\section{Introduction}
\label{sec:intro}

\let\thefootnote\relax\footnotetext{$^{\dag}$ Corresponding author}
The generation of a photorealistic representation, from an unseen point of view, also known as Novel View Synthesis, poses a long-standing challenge to the graphics and computer vision community. While there exist early methods that tackle this problem, both for static \cite{earlystatic1, earlystatic2} and dynamic scenes \cite{earlydynamic1,earlydynamic2}, neural rendering has received widespread attention only after the pioneering work of NeRF \cite{nerf}. Following the latter, numerous methods have contributed to the static neural rendering problem, by tackling aliasing problems \cite{mipnerf,mipnerf360}, generalizing to larger scales \cite{nerf++,blocknerf,zipnerf}, and to new scenes \cite{ibrnet} and increasing efficiency and rendering speed \cite{dvgo,plenoxels,instantngp,tensorf,factorfields}. 

\begin{center}
     \centering
     \captionsetup{type=figure}
     \includegraphics[width=\textwidth]{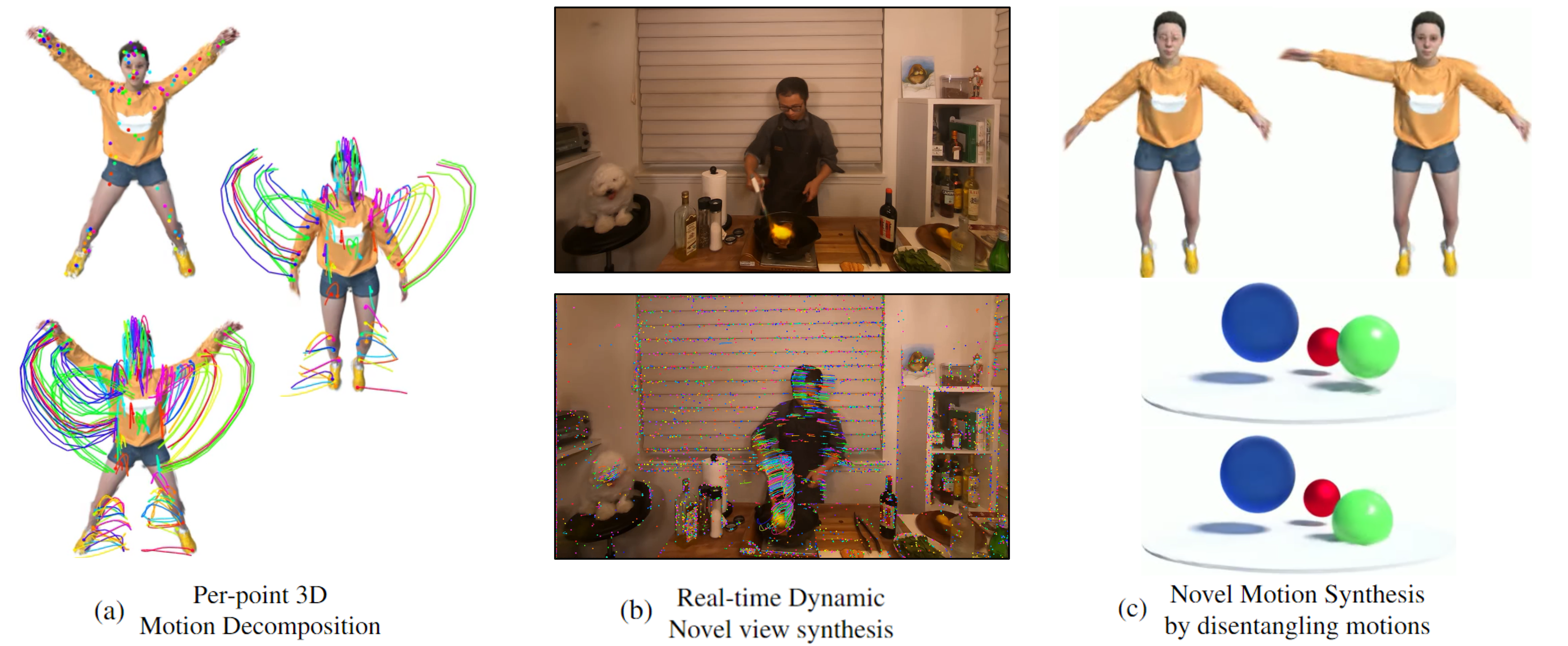}
     \captionof{figure}{Our framework can efficiently model all motions in a 3D dynamic scene (Figure 1a) through its neural motion decomposition scheme. We can synthesize photorealistic novel views, both for synthetic and real scenes in real-time, surpassing 120FPS for 1K resolution (Figure 1b). Our simple representation allows for novel motion synthesis by disentangling motions in the scene (Figure 1c). Notice how only the left hand or the green ball are moving in the `Jumpingjacks' and `Bouncingballs' D-NeRF scenes respectively. }
\end{center}

While the advance in novel view synthesis of static scenes has been rapid, the high-quality efficient photorealistic rendering of dynamic scenes still poses a vivid challenge to the aforementioned communities. The numerous applications, including but not limited to augmented reality/virtual reality (AR/VR), robotics, self-driving, content creation, and controllable 3D primitives for use in video games or the meta-verse, have kept the interest in expressive dynamic novel view synthesis to high levels. From recent pioneers of neural dynamic rendering \cite{dnerf,dynerf} to the latest \cite{3ddynamicgaussians}, numerous efforts have been made to drastically improve this field and bring it one step closer to real applications. Altering the scene representation from MLPs \cite{dnerf,dynerf} to grids \cite{mixvoxels}, or other decompositions \cite{kplanes,hyperreel,tensor4d,hexplane} for efficiency and faster training, proposing expressive deformation representations \cite{particlenerf,nerfplayer,hypernerf,nerfies,banmo} or using template guided methods \cite{humannerf1,humannerf2} are just a few of the approaches taken the last few years. 

Despite all these approaches, dynamic novel view synthesis is still a challenging problem, with quick training, real-time rendering, and highly accurate and visually appealing novel views yet to be achieved. Dynamic scenes require disambiguating displacements, and disentangling overlapping motions, especially due to non-rigid objects and monocular captured scenes that make this even more challenging. We propose a compact, simple, and expressive neural representation that can adequately tackle the previous challenges. Namely, we propose a sparse trajectory decomposition of all the motions in the dynamic scene, to accurately represent displacements, overlapping and non-rigid motions. Our framework is strategically designed to offer extremely fast training and real-time rendering of a dynamic scene. Specifically, we argue that a scene consists of only a few basis trajectories (even less than 10) and every point/motion shall be adequately mapped to one of them or a linear combination of them. The basis motions are learned through a tiny MLP queried only in time leading to superfast convergence during training (less than 5 minutes for photorealistic results, and half an hour for superior to the state-of-the-art ) and superfast rendering during inference (higher than 120 FPS for resolutions of 1K). By strategically constraining an inherently underconstrained problem in this way, we are able to successfully optimize for cases with very few correspondences (e.g. monocular videos). Simultaneously, being able to efficiently factorize all the motions of a dynamic scene into a few basis trajectories, allows us in turn, to control them, enable or disable them, opening a road to applications like video editing, scene controllability, interactive and real-time motion control, efficient animation and so on.   

In this work, we present \textbf{\DynMF} (\textbf{DYN}amic neural \textbf{M}otion \textbf{F}actorization). The contributions of this paper can be summarized as follows:
\begin{itemize}
    \item We manage to expressively model scene element deformation of complex dynamics, through a simple and interpretable framework. Our method enables robust per-point tracking, overcoming displacement ambiguities, overlappings, and non-rigid complex motions. It also enables the synthesis of novel motion scenes, that have never been seen before through disentangling motion and carefully enforcing sparsity. 
    \item \DynMF is extremely efficient. The carefully designed time-only queried MLP allows for training in less than 30 minutes while the rendering speed remains comparable to 3D Gaussian Splatting \cite{3dgs}. This allows for real-time rendering of static and dynamic scenes. 
    \item We propose a neural representation that can effectively constrain the ill-posed nature of dynamic 3D motion reconstruction and requires no prior knowledge of the scene. Our method achieves high-quality real-time dynamic rendering, while significantly surpassing the state-of-the-art benchmarks on the D-NeRF dataset and achieving comparable results on the real-scene datasets.
\end{itemize}

\section{Related Work}

\textbf{Dynamic Neural Rendering:} The recent works on NeRF \cite{nerf} and D-NeRF \cite{dnerf} both proposing a coordinate-based neural network, the former for querying points on a ray to produce color and density for static scenes and the latter to model the deformation field and create an expressive mapping between a canonical space and a dynamic scene, have provoked an explosion of interest in the field of static and dynamic novel view synthesis. On the dynamic side, we can distinguish three main approaches: (i) methods that construct a deformation field to warp a canonical space to every timestep \cite{dnerf,nerfies,hypernerf,trackingeea,banmo} (ii) methods that try to model the dynamic scene as a whole, implicitly or explicitly \cite{kplanes,hexplane,dynerf,mixvoxels,hyperreel} and (iii) point-based methods \cite{particlenerf,diffpointbased,3ddynamicgaussians} that due to dense correspondence over time between frames and due to their Lagrangian representation, can achieve consistent and expressive dynamic scene modeling. We relax that latter restriction by proposing a point-based method that can construct an efficient and expressive deformation field, explicit or learned, without the need for dense correspondences through multiple frames or multiple views.
\\
\\
\textbf{Motion Factorization:} Decomposing a complex motion field into a linear combination of basis trajectories has been a well-established approach from Mathematics and Physics to Signal Processing and Computer Vision. In the latter field, \cite{akh} has proposed an explicit trajectory-based factorization for non-rigid structure for motion in which every point in space can be linearly expressed by a combination of DCT trajectories. Clustering trajectories into a small number of subspaces has been argued by \cite{clustering} for the same task, while \cite{conv} has explored trajectories' convolutional structure. Following \cite{akh,otherlucy}, \cite{lucy} has also exploited DCT trajectory basis for the task of dynamic novel view synthesis, by predicting a per-point trajectory using an MLP to acquire the DCT coefficients. Our method argues that a per-point trajectory is redundant and can only work given a lot of correspondences or strong regularization. In fact, \cite{lucy} leverages optical flow in their optimization scheme. 
\\
\\
\textbf{Dynamic 3D Gaussians and Concurrent work:} During the past few months, numerous works have emerged to utilize 3D Gaussian Splatting \cite{3dgs} for Dynamic Neural Rendering.  
First, Dynamic 3D Gaussians \cite{3ddynamicgaussians} uses online training and rigidity losses to synthesize novel views timestamp per timestamp. For their method to work, though, they require multiple cameras and depth information to reconstruct the first frame, while at the same time, their method cannot handle the appearance of new objects in the scene. DynMF works on scenes captured by a single monocular camera and with random initialization incorporating information from all frames simultaneously. \cite{mlpgaussians} uses a coordinate-based MLP similarly to D-NeRF \cite{dnerf} as a deformation field. This approach may be slow and inefficient because it requires querying MLP for each Gaussian. Our method is extremely time-efficient and does not significantly depend on the number of Gaussians. \cite{4dgaussians} is optimizing 4D primitives to treat the spacetime as an entirety, while \cite{gaussian_hexplane} is connecting adjacent Gaussians via a HexPlane to model position and shape deformations. Our neural representation is simple and interpretable by decomposing all the complex motions of a dynamic scene into very few neural trajectories. Other concurrent works include \cite{efficient3dgr} that uses Fourier approximation and flow estimation, \cite{co-gs} which decomposes the scene's objects by designing a control-masking pipeline, while our method inherently enables the disentangling motion and controllability of the scene, and \cite{sc-gs} which uses local points that act as local bases to edit the deformable Gaussians. Other notable mentions are \cite{4dgsplatting,swags,mdsplatting,spacetimegaussians,gaufre,neuralparametric,control4d,gaussian-flow,3dgstream}.

\section{Method}

We propose a compact representation that is shared across all points and models the dynamics of a scene. We argue that every scene is comprised of a limited fixed number of trajectories. In other words, we can model a 3D dynamic scene by a few universal trajectories that are shared between the 3D world points. 
In this section, we first formally present the representation of our motion field, DynMF, which is able to model the 3D dynamic scene and synthesize novel views (Section \ref{subsection:mrf}). In Section \ref{subsection:3dgs} we review the framework in which we bundle our motion field representation to efficiently learn the 3D world, namely 3D Gaussian Splatting, while in Section \ref{subsection:optimization} we present the optimization and regularization details. 

\begin{figure}[t]
    \includegraphics[width=\linewidth]{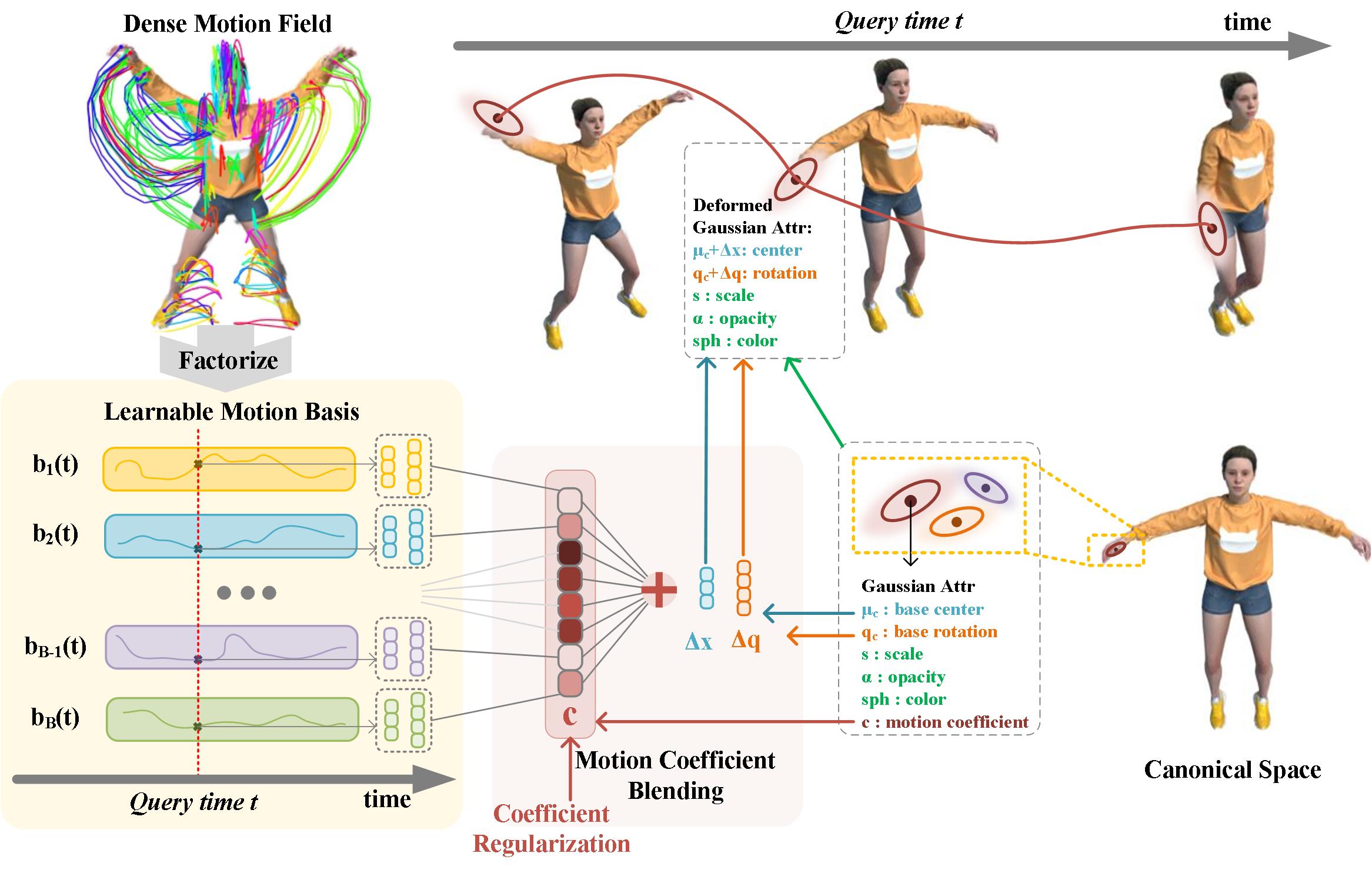}
    \captionof{figure}{Overview of \DynMF. The underlying dense motion field (left-top) of a dynamic scene is factorized into a set of globally shared learnable motion basis (left-bottom) and their motion coefficients stored on each Gaussian (right-bottom). Given a query time $t$, the deformation can be efficiently computed via a single global forward of the motion basis and the motion coefficient blending (middle-bottom) to recover the deformed scene (top-right).}
\vspace{-0.5cm}
\end{figure}

\subsection{Motion Representation Fields}
\label{subsection:mrf}
Let a dynamic scene be represented by a point cloud of N points $\bm{x} \in \mathbb{R}^{N\times3}$.
We assume a sequence of frames in the interval $t\in [0,T]$, taken from one or multiple viewpoints.
Assuming a canonical space $\bm{x_c} \in \mathbb{R}^{N\times3}$ we define $B$ trajectories $\bm{b}_j(t): [0,T]\rightarrow \mathbb{R}^3$ such that for every $\bm{x}\in \mathbb{R}^{1\times3}$ we have:
\begin{equation}
\label{eq:main}
\bm{x}(t) = \bm{x_c} + \sum_{j=1}^{B} c_j \bm{b}_j(t),
\end{equation}
where $c_j\in\mathbb{R}^{N\times 1}$ are the motion coefficients for each point, and $\bm{x}(t)$ are the points' motion trajectories across the entire sequence. We let the basis be shared across all points arguing that the dynamic motions of a scene can be explicitly represented by a linear combination of B trajectories. 
This factorization problem becomes an overconstrained low-rank factorization as soon as the number of points $N$ exceeds the number of basis trajectories $B$. 
This is easy to see if we discretize the time into $T$ timestamps and write only the x-coordinates as matrices $\bm{X},\bm{X_c} \in {\mathbb R}^{N\times T}$
\begin{equation}
\label{eq:main_matrix}
{\bm{X}}  = {\bm{X}_c} + \bm{C} \cdot \bm{B},
\end{equation}
with $\bm{C} \in {\mathbb R}^{N\times B}$ and $\bm{B} \in {\mathbb R}^{B\times T}$ and $B$ much lower than $N$ or $T$.

We reason that given enough correspondences and expressive basis functions, we can jointly learn the canonical space and each point's motion by reconstructing its trajectory across the whole sequence. 
Sharing the basis across all points enables the representation to produce physically plausible motions that stay consistent across all frames. 

If we let 
\begin{equation}
\label{eq:fourier}
\bm{b}_j(t)= \left\{
\begin{array}{ll}
      cos\left(\frac{2\pi j}{T}t\right), & \text{if j is odd} \\
      sin\left(\frac{2\pi j}{T}t\right), & \text{if j is even} \\
\end{array} 
\right.
\end{equation}
the factorization is equivalent to modeling the dynamic motion trajectories of a scene with Fourier series. While Fourier bases are global approximators for any trajectories and have been shown to be a compact representation for dynamic scenes with deformations  \cite{lucy1,lucy2}, we argue that a learned basis will be smoother and more expressive for this task (see Section \ref{sec:abl} for ablations). Thus, we let:
\begin{equation}
\label{eq:mlp}
    \mathbf{b}_j(t) = \textbf{MLP}\left( \frac{t}{T};w_j\right),
\end{equation}
be a small multilayer perceptron queried only over time. 
While the learned functions are not strictly a basis, 
 we empirically observe that they span a large enough space that covers the variation of deformations in the scene.
Overall, we propose a representation of the dynamic scene in which a small MLP predicts along all timestamps a few compact basis trajectories, and all the points in the scene are bound to linearly choose between these to model their motion across time.

\subsection{3D Gaussian Splatting}
\label{subsection:3dgs}

We choose to bind the aforementioned representation with 3D Gaussian Splatting \cite{3dgs}. The latter is an explicit 3D scene representation used for efficient and expressive neural rendering. 

\textbf{Static 3D Gaussians:} The 3D space is described in the form of a point cloud by a set of 3D Gaussians which are defined by a mean value $\bm{\mu}$ and a  3D covariance matrix $\bm{\varSigma}$ in the world space. The influence of one Gaussian with parameters $(\bm{\mu},\bm{\varSigma})$ on a given spatial position $\bm{x} \in \mathbb{R}^3$  is defined by:
\begin{equation}
    f(\bm{x}|\bm{\mu},\bm{\varSigma}) = e^{-\frac{1}{2}\left(\bm{x}-\bm{\mu}\right)^T \bm{\varSigma}^{-1}\left(\bm{x}-\bm{\mu}\right)}.
\end{equation}
The anisotropic full 3D covariance matrix is decomposed into a rotation matrix $\bm{R}$ and a scaling matrix $\bm{S}$:
\begin{equation}
    \bm{\varSigma} = \bm{R}\bm{S}\bm{S}^T\bm{R}^T.
\end{equation}
To render the 3D Gaussians in a differentiable manner \cite{3dgs} Gaussians are splatted  on the image plane by approximating the 2D means and covariances from the 3D ones.  The center of the Gaussian is splatted using the standard point rendering formula:
\begin{equation}
\label{eq:m3d}
\bm{\mu}^{2D} = \bm{K} \frac{\bm{E}\bm{\mu}}{\bm{E}\bm{\mu}_z},
\end{equation}
where $K, E$ are the intrinsic projection matrix and the world-to-camera extrinsic matrix, respectively. The 3D covariance matrix is splatted into 2D using \cite{zwicker2002ewa}:
\begin{equation}
\bm{\varSigma}^{2D} = \bm{J}\bm{E}\bm{\varSigma}\bm{E}^T\bm{J}^T,
\end{equation}
where $\bm{J}$ is the Jacobian of the point projection formula in~\ref{eq:m3d}.
Finally, the color of a specific pixel $p$ can be computed as:
\begin{equation}
    C_p = \sum_{i=1}^{N} c_i \alpha_i f\left(p|\bm{\mu}^{2D},\bm{\varSigma}^{2D} \right)\prod_{j=1}^{i-1} \left(1-\alpha_j f\left(p|\bm{\mu}^{2D},\bm{\varSigma}^{2D}\right) \right),
\end{equation}
where $c_i,\alpha_i$ are the color and opacity, bound with each 3D Gaussian.
Overall, each Gaussian is characterized by the following attributes: i. a center mean $\bm{\mu} \in \mathbb{R}^3$, ii. a scale factor $\bm{s} \in \mathbb{R}^3 $ and a rotation $\bm{R} \in SO(3)$ that we parametrize with a unit quaternion in ${\mathbb S}^3$, iii. an opacity logit $\alpha\in\mathbb{R}$ and iv. each color defined by spherical harmonics coefficients $\bm{c}_{R,G,B}\in\mathbb{R}^{(L+1)^2}$ where $L$ is the maximum degree of spherical harmonics.

\textbf{Dynamic 3D Gaussians:} The previous framework is capable of modeling efficiently and expressively a 3D static scene. To use it for representing a 3D dynamic scene we propose the following:
\begin{enumerate}
  \item Substituting the mean $\bm{\mu}$ with $\bm{\mu}(t)$. The mean's trajectory $\bm{\mu}(t)$ will be decomposed according to Equations \ref{eq:main} and \ref{eq:main_matrix}:
  \begin{equation}
      \bm{\mu}(t) = \bm{\mu}_c + \sum_{j=1}^B c_j \bm{b}^\mu_j(t),
  \end{equation}
  where $\bm{\mu}_c\in\mathbb{R}^3$ is the mean center of the Gaussian in the canonical space and $\bm{c} = \left[c_j\right]_{j=1,\dots, B}$ the motion coefficients, both stored as learned parameters per Gaussian. 
  
  \item  Substituting the rotation quaternion $\bm{q}$ with $\bm{q}(t)$. The rotation quaternion will be decomposed similarly:
    \begin{equation}
      \bm{q}(t) = \bm{q}_c + \sum_{j=1}^B c_j \bm{b}^R_j(t),
    \end{equation}
  where $\bm{q}_c\in\mathbb{R}^4$ is the rotation quaternion of the Gaussian in the canonical space stored as learned parameter per Gaussian. Note that the motion coefficients are shared for both the translational and rotational decompositions. 
  \item Keeping density and spherical harmonic coefficients of each Gaussian fixed along time. The basis functions $\{\bm{b}_j^\mu(t)\}$ and $\{\bm{b}_j^R(t)\}$ are chosen according to Equations \ref{eq:fourier} and \ref{eq:mlp}.
\end{enumerate}

We argue that the proposed framework is a compact and expressive representation of 3D dynamic scenes that consist of multiple decoupled and complicated motions. By sharing the basis functions between Gaussians, we constrain neighbor Gaussians and rigid areas to choose similar trajectories, softly enforcing locality and rigidity a priori. At the same time, sharing the motion coefficients and keeping the number of basis trajectories low, dramatically softens the restrains of this ill-posed optimization problem.  The efficient design of the basis functions retains the rendering speed at the same levels as with the original Gaussian Splatting, enabling real-time rendering not only for static scenes but for dynamic as well.

\subsection{Optimization Framework}
\label{subsection:optimization}
\label{sec:opt}
\textbf{Regularization:} The optimization process will naturally converge to a linear combination of all basis functions per Gaussian. Furthermore, Gaussians tend to move, even if they are part of the background as long as they do not harm the photometric accuracy with their movement (e.g. Gaussians moving on a white wall). While the former is desirable and increases expressivity, the latter harms the representation's interpretability since the basis functions are forced to learn trajectories for unnecessary movements. To solve this problem, we add an L1 regularization loss on the motion coefficients to penalize unnecessary choices of basis functions and thus, apparent but nonexistent movements:
\begin{equation}
    \mathcal{L}_1 = \frac{1}{NB}\sum_{i=1}^N \sum_{j=1}^B |c_{ij}|.
\end{equation}
\\
\textbf{Sparsity:} Enforcing sparsity between the trajectories is more challenging. While $\mathcal{L}_1$ loss is supposed to encourage sparsity by increasing competition between the weights, we observed that the moving Gaussians kept choosing a linear combination of the trajectories with just smaller weights. To solve this we added a new stronger loss that explicitly forces each Gaussian to choose only very few trajectories:
\begin{equation}
\mathcal{L}_s = \frac{1}{N} \sum_{i=1}^{N} \frac{\left \lVert c_i \right \rVert_1}{\left \lVert c_i \right \rVert_\infty}.
\end{equation}
With a strong enough loss weight, this quantity enforces strict sparsity which can lead to dynamic motion decoupling and novel motion synthesis. 
\\
\\
\textbf{Rigidity:} Our representation naturally encourages rigidity between the Gaussians. Restraining the number of trajectories drives nearby points to choose the same trajectory. Furthermore, when a Gaussian is split or densified into a new one, we choose to copy the motion coefficients to the new Gaussian, to naturally enforce rigidity to the scene. We experiment with a simple rigid loss:
\begin{equation}
\mathcal{L}_r = \frac{1}{Nk}  \sum_{i=1}^N \sum_{j\in knn_{i;k}} w_{ij}\lVert \bm{c}_i - \bm{c}_j\rVert_2,
\end{equation}
where the set of $j$ Gaussians choose the k-nearest neighbours of $i$ and weight the loss, similar to \cite{3ddynamicgaussians}, by  an isotropic Gaussian weighting factor per Gaussian pair $w_{ij} = exp\left(-\lambda_w \lVert \bm{\mu}_i(t^*)- \bm{\mu}_j(t^*)\rVert_2^2 \right)$.
We demonstrate that $\mathcal{L}_r$ does not further increase the rendering quality of the scene, solidifying our intuition that our motion representation is inherently rigid, given a small number of trajectories. For ablations on the optimization scheme, please refer to the Supplementary material.

\begin{figure}
    \includegraphics[width=\linewidth]{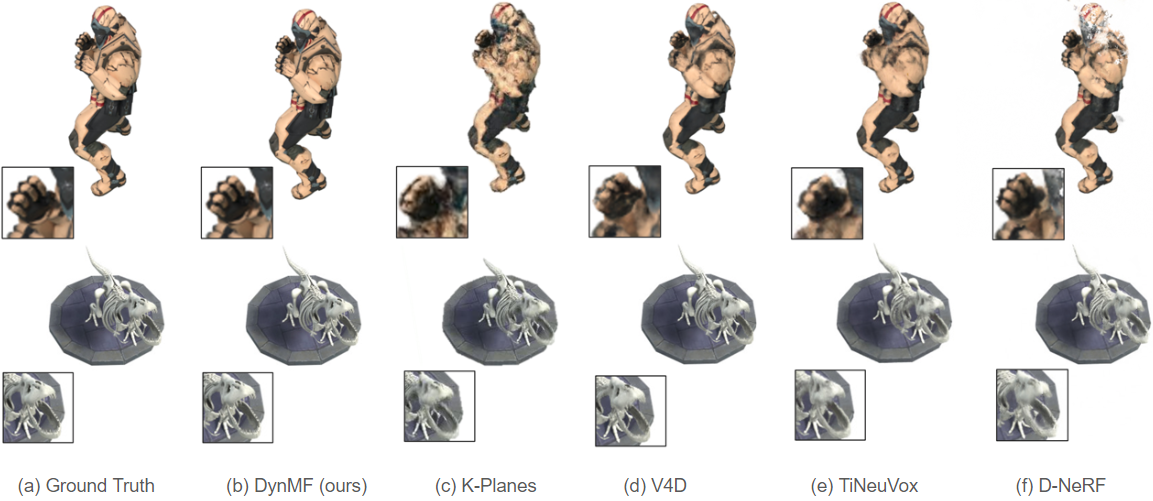}
    \captionof{figure}{Qualitative comparison on the D-NeRF dataset between our method and the state-of-the-art.}
    \label{fig:dnerf}
\vspace{-0.5cm}
\end{figure}

\begin{table}
\caption{Quantitative results on synthetic monocular D-NeRF dataset. The rendering resolution is 400×400. \textsuperscript{1} Concurrent work that utilizes Gaussian-splatting. \textsuperscript{2} Rendering resolution is 800$\times$800.}
\centering
\scalebox{1}{
\begin{tabular}{c|ccc|c|c} \hline 
Model & PSNR(dB) $\uparrow$ & SSIM $\uparrow$ & LPIPS $\downarrow$ & Time $\downarrow$ & FPS $\uparrow$  \\ \hline 
D-NeRF \cite{dnerf}  & 29.67 & 0.953 & 0.06 & $15.9\mathrm{hours}$ & $ 0.1$  \\
K-Planes \cite{kplanes} & 31.61 & 0.970 & $-$ & $52 \mathrm{mins}$ & 0.1  \\
TiNeuVox \cite{tineuvox} & 32.67 & 0.970 & 0.04 & $\mathbf{\mathrm{21m}}$ & 1.5  \\
V4D \cite{v4d} & 33.72 & 0.980 & -- & $4.9 \mathrm{hours}$ & $2$\\ \hline
\cite{efficient3dgr}\textsuperscript{1} & 32.07 & 0.96 & -- & $8 \mathrm{mins}$ & --\\ 
\cite{4dgsplatting}\textsuperscript{1} & 34.26 & 0.97 & 0.03 & $\mathbf{5} \mathbf{mins}$ & $1257$\\
Gaufre \cite{gaufre}\textsuperscript{1} & 34.80 & 0.982 & 0.02 & $25 \mathrm{mins}$ & 50 \\
4DGS \cite{gaussian_hexplane}\textsuperscript{1,2}  & 34.05 & 0.98 & 0.02 & $20 \mathrm{mins}$ & 82 \\ \hline
DynMF (Ours) & $\mathbf{3 6 . 89}$ & $\mathbf{0 . 9 83}$ & $\mathbf{0 . 0 2}$ & $20 \mathrm{mins}$ & $\mathbf{300}$ \\ \hline
\end{tabular}
}

\label{table:dnerf}
\end{table}

\section{Experiments}
\label{sec:experiments}

\subsection{Experimental Setup}
In this section, we present the results of our method for three different scenarios; synthetic-scene monocular, real-scene monocular and real-scene multi-view. 

\textbf{D-NeRF} \cite{dnerf} is a monocular synthetic dataset with a total of 8 rigid and non-rigid scenes. Each scene contains dynamic frames, ranging from 50 to 200 in number and on a unique timestamp. The camera pose for each timestamp is also different. 
\textbf{HyperNeRF} \cite{hypernerf} is a monocular dataset on real-scenes. It consists of many scenes of different real objects and actions with one or two cameras that are moving over time. We choose one rigid and three highly non-rigid scenes to test the method's efficiency in challenging cases like these. The four scenes are the `3d-printer', `chicken', `broom', and `peel-banana'. 
\textbf{N3DV dataset} \cite{dynerf} dataset is a real-world dataset consisting of 6 very challenging dynamic scenes. The scenes are captured by 15-20 static cameras in total that share a common time. The scenes consist of long highly non-rigid motions and dynamic opacity (e.g. flame or smoke).

\subsection{Implementation Details}
The motion representation field is implemented in PyTorch \cite{pytorch}, while we keep the differentiable Gaussian rasterization implemented by 3D-GS \cite{3dgs}. The MLP consists of 8 linear layers of 256-width with 2 final layers producing $(B,3)$ and $(B,4)$ numbers that constitute the displacement and quaternion correction per trajectory for a specific time. For the number of trajectories $B$, we choose $10$ for D-NeRF and $32$ for HyperNeRF and N3DV dataset. We apply positional encoding to the input of the MLP, which has been shown to be fundamental for good performance \cite{nerf,penerf}. Since the network is not coordinate-based and is solely based on time, a high-frequency input is needed for producing expressive trajectories for the dynamic scenes. We choose 32 frequencies to encode time. See Section \ref{sec:abl} for ablations on the number of trajectories and position encoding of time. For training, we conduct training for a total of 30k iterations, while enabling the training of the basis functions and motion coefficients only after the first 3k iterations. All the parameters are kept the same with respect to 3D-GS while the learning rate of the basis is exponentially decaying from 8e-4 to 8e-6 and the learning rate of the coefficients is kept at 8e-3. 2k Gaussians' means are randomly initialized around the center of the scene for all datasets, solidifying the stability and generalizability of our representation.

\begin{figure}[t]
    \centering
    \scalebox{0.8}{
    \includegraphics[width=\linewidth]{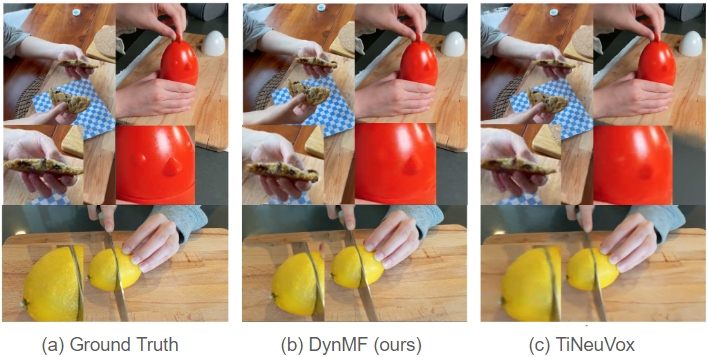}}
    \captionof{figure}{Qualitative comparison on the HyperNeRF dataset.}
    \vspace{-0.3cm}
\end{figure}
\begin{table}[t]
\setlength{\tabcolsep}{3pt}
\centering
\caption{Per-scene quantitative comparisons on HyperNeRF monocular real dynamic scenes.  \textsuperscript{1} Concurrent work that utilizes Gaussian-splatting. }
\scalebox{0.9}{
\begin{tabular}{ccccccccccccc}
\toprule
& \multicolumn{2}{c}{3D Printer} & \multicolumn{2}{c}{Chicken} & \multicolumn{2}{c}{Broom} & \multicolumn{2}{c}{Peel Banana} \\
\textbf{Method} 
 & PSNR$\uparrow$  & SSIM$\uparrow$ 
& PSNR$\uparrow$  & SSIM$\uparrow$ 
 & PSNR$\uparrow$  & SSIM$\uparrow$
 & PSNR$\uparrow$  & SSIM$\uparrow$ \\
\cmidrule(lr){1-1}  \cmidrule(lr){2-3} \cmidrule(lr){4-5} \cmidrule(lr){6-7} \cmidrule(lr){8-9}
    Nerfies~\cite{nerfies}
     & 20.6 & 0.83 
     & 26.7 & 0.94 
     & 19.2 & 0.56 
     & 22.4 & 0.87   \\
    HyperNeRF~\cite{hypernerf}
     & 20.0 & 0.59 
     & 26.9 & 0.94
     & 19.3 & 0.59 
     & 23.3 & 0.90   \\
    TiNeuVox~\cite{tineuvox}
    & \textbf{22.8} & \textbf{0.84}
    & 28.3 & \textbf{0.95}
    &  21.5 &  0.69
    & 24.4 & 0.87  \\ \hline
    3D-GS~\cite{3dgs}\textsuperscript{1}
    & 18.3 & 0.60
    & 19.7 & 0.70
    &  20.6 &  0.63
    & 20.4 & 0.80  \\
    4DGS~\cite{hexplane}\textsuperscript{1}
    & 22.1 & 0.81
    & 28.7 & 0.93
    & 22.0 &  \textbf{0.70}
    & 28.0 & \textbf{0.94}  \\
    ~\cite{4dgsplatting}\textsuperscript{1}
    & 21.1 & 0.75
    & 25.8 & 0.86
    & 20.4 &  0.66
    & 26.6 & 0.92  \\ \hline
    DynMF (ours)
    &  22.4  & 0.70
    &  \textbf{28.5} & 0.81
    &  \textbf{22.1} & 0.61
    & \textbf{28.1}& 0.89  \\
    \bottomrule
\end{tabular}
}
\label{table:hypernerf}
\vspace{-0.7cm}
\end{table}

\begin{figure}[ht]
    \includegraphics[width=\linewidth]{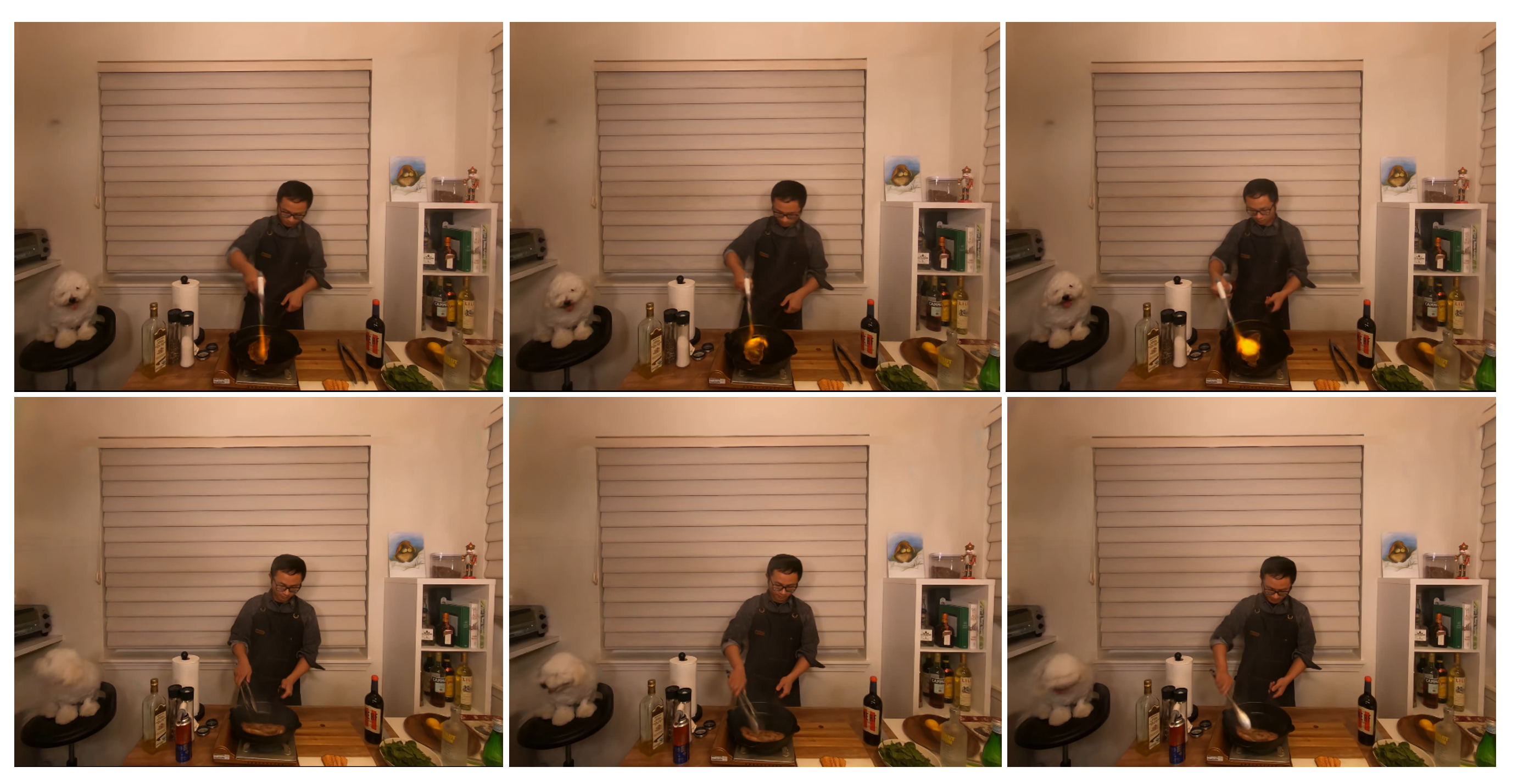}
    \captionof{figure}{Novel view synthesis of DynMF. The figure shows selected frames from different time frames for the scenes 'flame-steak' and 'sear-steak'.}
    \vspace{-0.3cm}
\end{figure}

\begin{figure}
    \includegraphics[width=\linewidth]{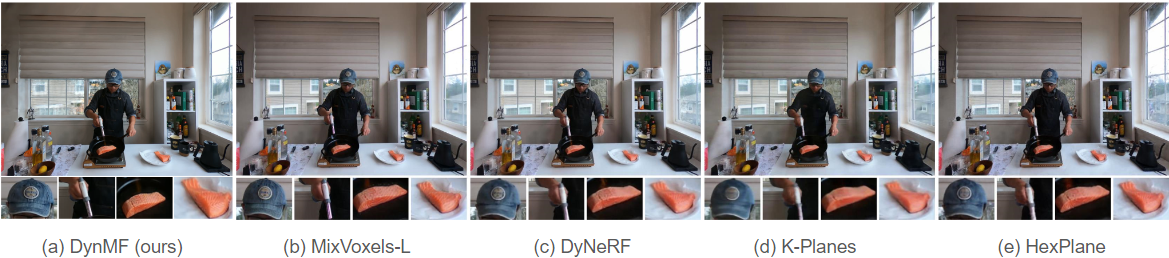}
    \captionof{figure}{Qualitative comparison on the N3DV dataset between our method and the state-of-the-art.}
    \vspace{-0.5cm}
\end{figure}

\begin{table}
\caption{Quantitative results on real-scene multi-view N3DV dataset. The rendering resolution is $1386\times1014$. \textsuperscript{1} Concurrent work that utilizes Gaussian-splatting.}
\centering
\scalebox{1}{
\begin{tabular}{c|ccc|c|c}

\hline Model & PSNR(dB) $\uparrow$ & SSIM $\uparrow$ & LPIPS $\downarrow$ & Time $\downarrow$ & FPS $\uparrow$  \\ \hline 

DyNeRF \cite{dynerf} & 29.58 & - & $\mathbf{0.08}$ & $1344\mathrm{hours}$ & $0.015$  \\
MixVoxels-L \cite{mixvoxels}   &30.80 & 0.96 & - & $1.3 \mathrm{hours}$ & $16.7$   \\
K-Planes \cite{kplanes}& 31.63 & 0.964 & - & $1.8 \mathrm{hours}$ & 0.15  \\
Hexplane \cite{hexplane} & $\mathbf{32.22}$ &  $\mathbf{0.98}$ & 0.09 & $12 \mathrm{hours}$ & 0.09\\ \hline
\cite{efficient3dgr}\textsuperscript{1} &  28.89 & 0.945 & - & $1 \mathrm{hours}$ & 118\\
\cite{4dgsplatting}\textsuperscript{1} & 31.62 & 0.94 & 0.14 & $1 \mathrm{hours}$ & \textbf{277}\\
4DGS \cite{4dgaussians}\textsuperscript{1} & 32.01 &  - &  \textbf{0.06} & - & 114\\ \hline
DynMF (Ours) & 31.70 & 0.946 & 0.18 & $\mathbf{40mins}$ & 135 \\
\hline
\end{tabular}
}
\label{table:dynerf}
\end{table}

\subsection{Results}
\textbf{Results on Synthetic data:} A quantitative evaluation of the D-NeRF synthetic Dataset is presented in Table \ref{table:dnerf}. Our method surpasses all state-of-the-art methods by a significant gap in all metrics. This confirms that DynMF is an adequate representation of rigid, decoupled, and non-rigid/deformable motion as well. Regarding speed, our method achieves state-of-the-art results in only 10 minutes of training, while after half an hour it synthesizes most of the dynamic scenes almost perfectly. The FPS rate for $400\times400$ images reaches more than 300FPS confirming the carefully designed representation in terms of efficiency, achieving real-time dynamic novel view synthesis. Qualitative comparisons can be seen in Figure \ref{fig:dnerf}.
\\
\\
\textbf{Results on Real data:} DynMF seems to achieve state-of-the-art results on the HyperNeRF and N3DV datasets as well, as shown in Tables \ref{table:hypernerf} and \ref{table:dynerf}. These datasets consist of very challenging non-rigid motions and semi-transparent areas that the representation of motion factorization can adequately and efficiently express. Again, time-wise our method allows for real-time dynamic rendering even on scenes of high resolution like N3DV (i.e. $1280\times1080$). For a per-scene evaluation of all three datasets and for more qualitative results, please refer to the supplementary material. 

\begin{figure}
    \scalebox{1}{\includegraphics[width=\linewidth]{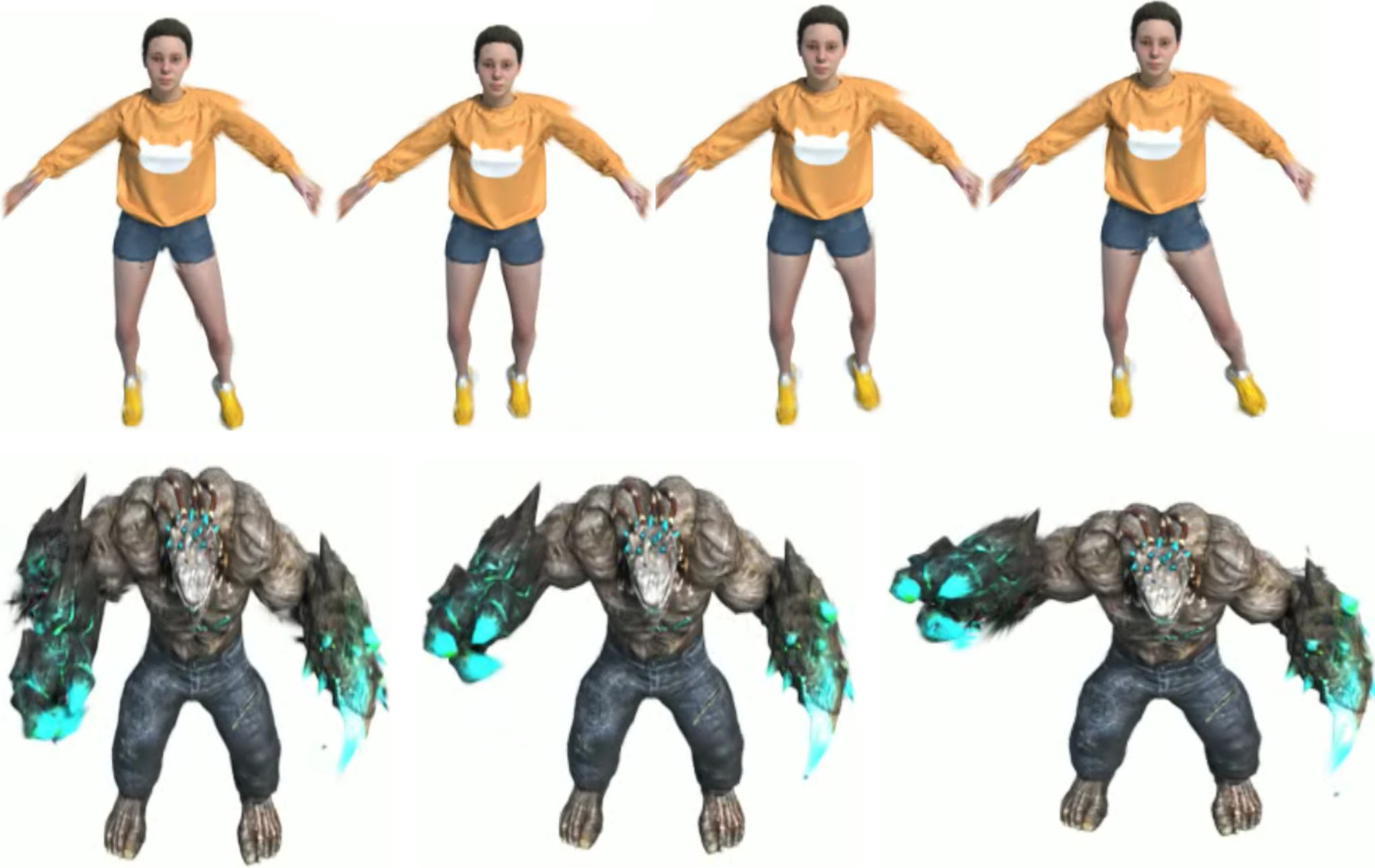}}
    \captionof{figure}{Demonstration of the decomposition capabilities of our representation. Independent movement of the right foot and the left hand of the `Jumpingjacks' and `Mutant' scenes respectively.}
\label{fig:decomposition}
\end{figure}

\begin{table}
\vspace{-0.3cm}
\caption{Ablation study on the number $B$ of the motion trajectories and the frequencies $F$ of the positional encoding applied on time. PSNR is used as the metric on the `Mutant' scene from the D-NeRF dataset and the `Flame-steak' scene from the N3DV dataset.}
\centering
\scalebox{1}{
\begin{tabular}{c|cc||c|cc}
\hline Model (B, F) & Mutant & Steak &  Model (B, F) & Mutant & Steak\\
\hline 
DynMF (1,32) & 29.75 & 28.85 & DynMF (10,1) & 39.08 & 29.32 \\
DynMF (4,32) & 39.98 & 30.34 & DynMF (10,6) & 39.61 & 29.59 \\
DynMF (10,32) & \textbf{40.79} & \textbf{31.57} & DynMF (10,16) & 40.50 & 31.29  \\
DynMF (20,32) & 40.76 & 31.03 & DynMF (10,32) & \textbf{40.79} & \textbf{31.57} \\
DynMF (50,32) & 40.53 & 30.99 & DynMF (10,60) & 40.53 & 29.80\\
\hline
\end{tabular}
}
\label{table:abl}
\end{table}

\subsection{Motion Decomposition}
A key component of our proposed representation is its ability to explicitly decompose each dynamic scene to its core independent motions. Specifically, by applying the sparsity loss described in Section \ref{sec:opt}, we enforce each Gaussian to choose only one of the few trajectories available. This design combined with the inherent rigid property of our representation drives all nearby Gaussians to choose the same and only trajectory. This strongly increases the controllability of the dynamic scene, by disentangling motions, allowing for novel scene creation, interactively choosing which part of the scene is moving, and so on. Figure \ref{fig:decomposition} shows two examples of motion control in two synthetic scenes. Please, refer to the supplementary video for a deeper understanding of the motion decomposition possibilities that our representation enables.

\vspace{-0.3cm}
\subsection{Ablation Studies}
\label{sec:abl}

\textbf{Time encoding and Number of trajectories:} The number of trajectories, $B$ is the most important hyperparameter of DynMF, for it is the one that decides in what detail will the scene be modeled. In parallel, the positional encoding on time proved to be important as well, with notoriously high frequencies, $F$, needed for state-of-the-art results. Table \ref{table:abl} shows experiments on multiple combinations of $(F, B)$. We can see that both synthetic and real scenes can be adequately modeled by very few trajectories while adding more only subtly increases the performance. At the same time, while encoding time is not vital for satisfactory rendering, mapping it to higher frequencies provides the mean for more expressive trajectories by the small MLP. Please, refer to the Supplementary material for qualitative comparisons for different values of $B$ and $F$. 
\newline
\newline
\textbf{Fourier Series:} Table \ref{table:fourier} shows results on two different scenes (mutant from D-NeRF and flame-steak from DyNeRF) in the case of a non-learned motion and rotation base, in this case the Fourier Series. We conducted a few experiments with the basis functions following Equation \ref{eq:fourier} and we varied the number of frequencies the basis could incorporate. We observe that the Fourier series can satisfactorily express the motion field of a dynamic scene but is not expressive and smooth enough to decode the details of complicated scenes. To the best of our knowledge, this is the first totally explicit dynamic scene representation, comprised of Gaussians and Fourier series. In supplementary material, we further investigate the reasons why the Fourier basis is not expressive enough. 

\begin{table}
\vspace{-0.6cm}
\caption{Ablation study on the basis function. PSNR is used as the metric on the ‘Mutant’ scene from the
D-NeRF dataset and the ‘Flame-steak’ scene from the N3DV dataset. }
\centering
\scalebox{0.9}{
\begin{tabular}{l|ccc|c|c}
\hline Model & Mutant & Steak \\
\hline DynMF (Learned Basis) & \textbf{40.79} & \textbf{31.57}\\
DynMF (Fourier 10) & 33.17 & 28.95   \\
DynMF (Fourier 20) & 35.21 & 27.18   \\
DynMF (Fourier 50) & 33.87 & 23.42\\
DynMF (Fourier 100) & 21.12 & 19.15 \\
\hline
\end{tabular}
}
\label{table:fourier}
\vspace{-1cm}
\end{table}

\section{Conclusion \& Future Work}

In this work, we have introduced, \DynMF, a motion factorization framework that enables real-time rendering of dynamic scenes. The carefully designed representation of a small MLP queried only by time allows for training convergence in half an hour and for a rendering speed of more than 120 FPS. At the same time, the shared motion basis scheme supports inherently the locality and rigidity of the motions and allows for state-of-the-art rendering quality from novel views. Last but not least, the motion decomposition properties allow us to dive deeper into the structure of the scene and create novel instances by independently enabling and disabling motions on the scene. For future work, we believe that \DynMF is potentially a good representation for efficient and accurate tracking as well. While the motion factorization framework along with the 3D Gaussian splatting gives the opportunity for detailed and expressive novel view synthesis of dynamic scenes, it is by itself constructed in a way to allow tracking of all the points on the scene at every time.

\clearpage

\textbf{Acknowledgment:} The authors appreciate the support
of the gift from AWS AI to Penn Engineering’s ASSET
Center for Trustworthy AI; and the support of the following
grants: NSF IIS-RI 2212433, NSF FRR 2220868 and ONR N00014-22-1-2677 awarded
to UPenn

\bibliographystyle{splncs04}
\bibliography{main}

%
%
\end{document}